\documentclass[11pt]{article}

\PassOptionsToPackage{hyperfootnotes=false}{hyperref}

\usepackage[final]{acl}

\usepackage{times}
\usepackage{latexsym}

\usepackage[T1]{fontenc}

\usepackage[utf8]{inputenc}

\usepackage{microtype}

\usepackage{inconsolata}

\usepackage{graphicx}

\usepackage{booktabs}
\usepackage{siunitx} 
\usepackage{adjustbox}
\usepackage{multirow}
\usepackage{threeparttable}

\usepackage{amsfonts}
\usepackage{amsmath}

\usepackage[]{mdframed}
\usepackage{fancyvrb}
\usepackage{fvextra}

\usepackage{array}
\usepackage{rotating}
\usepackage{subcaption}

\usepackage{pgfplots}
\pgfplotsset{compat=1.17}

\definecolor{correct}{HTML}{036400}
\definecolor{wrong}{HTML}{CB0000}

\newcommand{\Ni}{({\em i})~}
\newcommand{\Nii}{({\em ii})~}
\newcommand{\Niii}{({\em iii})~}

\title{Redefining Retrieval Evaluation in the Era of LLMs}

\author{
\textbf{Giovanni Trappolini\textsuperscript{1}}\thanks{These authors contributed equally to this work.}\thanks{\texttt{trappolini@diag.uniroma1.it}}, 
\textbf{Florin Cuconasu\textsuperscript{1,2}$^*$}\thanks{Work conducted while FC being a research intern at TII.}\thanks{\texttt{cuconasu@diag.uniroma1.it}}\\
 \textbf{Simone Filice\textsuperscript{2},}
 \textbf{Yoelle Maarek\textsuperscript{2},}
 \textbf{Fabrizio Silvestri\textsuperscript{1}}
\\
\\
 \textsuperscript{1}Sapienza University of Rome,
 \textsuperscript{2}Technology Innovation Institute
}

\begin{document}
\maketitle
\begin{abstract}

Traditional Information Retrieval (IR) metrics, such as nDCG, MAP, and MRR, assume that human users sequentially examine documents with diminishing attention to lower ranks. This assumption breaks down in Retrieval Augmented Generation (RAG) systems, where search results are consumed by Large Language Models (LLMs), which, unlike humans, process all retrieved documents as a whole rather than sequentially. 
Additionally, traditional IR metrics do not account for related but irrelevant documents that actively degrade generation quality, rather than merely being ignored. Due to these two major misalignments, namely human vs. machine position discount and human relevance vs. machine utility, classical IR metrics do not accurately predict RAG performance.
We introduce a utility-based annotation schema that quantifies both the positive contribution of relevant passages and the negative impact of distracting ones. Building on this foundation, we propose UDCG (Utility and Distraction-aware Cumulative Gain), a metric using an LLM-oriented positional discount to directly optimize the correlation with the end-to-end answer accuracy.
Experiments on five datasets and six LLMs demonstrate that UDCG improves correlation by up to 36\% compared to traditional metrics.
Our work provides a critical step toward aligning IR evaluation with LLM consumers and enables more reliable assessment of RAG components\footnote{Code and data at \href{https://github.com/GiovanniTRA/UDCG}{https://github.com/GiovanniTRA/UDCG}}.
\end{abstract}

\section{Introduction}\label{sec:intro}

The emergence of Retrieval Augmented Generation (RAG) systems has fundamentally transformed how we approach knowledge-intensive natural language processing tasks \cite{gao2024retrievalaugmentedgenerationlargelanguage}. By combining the parametric knowledge of Large Language Models (LLMs) with the dynamic retrieval of relevant information from external corpora, RAG systems have demonstrated remarkable capabilities across diverse applications, including question answering \cite{lewis2020retrieval,izacard-grave-2021-leveraging} and fact-checking \cite{khaliq-etal-2024-ragar} and beyond. However, as these systems have gained prominence, a critical gap has emerged between the evaluation methodologies inherited from traditional information retrieval (IR) and the unique characteristics of LLM-based retrieval consumers.

Traditional offline IR evaluation metrics, including Normalized Discounted Cumulative Gain (nDCG) \cite{10.1145/582415.582418}, Mean Average Precision (MAP), and Mean Reciprocal Rank (MRR), were developed with human users in mind \cite{manning2008IR}. In other words, they assume that the ultimate judge of the quality of the results is a human. 
These assumptions cause two major limitations in the generative AI and agentic AI eras, when LLMs or more generally {\em machines} consume the search results. 

\noindent \textbf{Human vs. Machine Position Discount.} 
Humans sequentially examine retrieved documents; therefore, human-centric IR metrics assign monotonically decreasing weights to lower-ranked documents, reflecting the intuitive notion that documents appearing further down the list are less likely to be examined and therefore less valuable to the user. 
Conversely, in the case of RAG systems, LLMs process all retrieved documents as a whole within their input context.
%
%
Recent research has revealed that LLMs exhibit systematic positional biases when processing prompts that include multiple documents \cite{liu-etal-2024-lost, hutter-2025-lost-but-not-only}. These biases can manifest in various patterns depending on the model and context, potentially causing LLMs to attend differently to information based on its position in the prompt. This suggests that the monotonic position-based discount embedded in classical IR metrics may not accurately reflect how LLMs leverage retrieved content, and how it should be valued.

\noindent \textbf{Human Relevance vs. Machine Utility.}
A second limitation is that traditional IR metrics operate under a simplistic binary (relevant vs. irrelevant) or ordinal (e.g., 0-5 relevance scale) relevance framework, treating all non-relevant documents as equivalent. This assumption fails to capture two crucial aspects of the RAG setting: \Ni LLMs might fail to use relevant passages (e.g., because they are convoluted); and \Nii irrelevant documents not only do not contribute value, they actually have a negative impact as they act as distractors that can mislead the LLM and degrade the system performance \cite{cuconasu2024pon,jin2025longcontext}. The heterogeneous nature of these distracting effects means that different irrelevant documents can have vastly different impacts on the quality of the final generation \cite{amiraz2025distracting}. Yet, current metrics provide no mechanism to distinguish between them.

These two limitations of existing IR relevance metrics extend beyond theoretical concerns. They have practical implications for RAG system development and deployment, since IR components are optimized using these metrics, which may actually harm end-to-end system performance. We argue here that RAG systems face today a fundamental challenge: \emph{how can they be reliably evaluated and improved, when their retrieval component is optimized for metrics designed for an entirely different use case?} This paper addresses this challenge through both empirical analysis and methodological innovation.

We begin by providing concrete demonstrations of how traditional IR metrics fail to predict RAG performance, using constructed examples that highlight the disconnect between metric scores and actual system effectiveness.  To address this disconnect, we propose a novel passage annotation strategy that fundamentally reconceptualizes relevance assessment for LLM consumers. Rather than relying on discrete relevance judgments designed for human interpretation, we introduce a continuous utility scoring framework that captures both the beneficial impact of relevant passages on LLM generation quality and the distracting effects of irrelevant passages that can mislead the model. This dual consideration of utility and distraction enables a more nuanced understanding of how individual passages contribute to or detract from end-to-end system performance.
We demonstrate the practical value of this annotation approach by showing that ranking retrieved passages according to our utility scores and selecting the top-$k$ results yields significantly better LLM accuracy compared to selection based on traditional relevance annotations. This finding provides empirical evidence that optimizing for utility-aware rankings directly translates to improved RAG system performance.

Building on these utility-based annotations, we introduce a novel evaluation metric specifically designed for RAG systems that we refer to as Utility and Distraction-aware Cumulative Gain (UDCG). UDCG leverages our continuous scoring framework to substitute human relevance with machine utility. Moreover, it applies an LLM-oriented positional discount that directly optimizes the correlation between the IR metric and the end-to-end answer accuracy. 
Through comprehensive experiments across six different LLMs and five diverse benchmarks, we demonstrate that UDCG exhibits substantially stronger correlation with end-to-end RAG accuracy than traditional IR metrics, with increments up to 36\%. These results establish UDCG as a more reliable indicator of retrieval component quality for RAG systems, enabling practitioners to optimize their systems toward metrics that actually predict downstream performance.

\section{Related Work}\label{sec:related}

\subsection{Retrieval Augmented Generation}\label{sec:related_rag}

Recently, RAG has emerged as a new application of IR systems: LLMs by accessing information retrieved with IR systems can incorporate external knowledge into their outputs, improving the response quality \cite{lewis2020retrieval,gao2024retrievalaugmentedgenerationlargelanguage,fan2024survey}. This approach addresses critical limitations of standalone language models, particularly their tendency toward hallucination and their inability to access information beyond their training cutoff dates. 
When operating in the RAG setting, LLMs have two major limitations related to their ability to capitalize on the input context. 
First, the effect of irrelevant documents, defined in \citet{cuconasu2024pon} as documents not containing the answer to the question. Humans are not really confused by their presence in the result list since they can quickly skim and dismiss them.
On the contrary, for LLMs in RAG systems, irrelevant documents pose a more serious problem: they act as noise that can dilute attention, trigger hallucinations, or cause the model to synthesize incorrect information by attempting to find connections between the query and unrelated content. \citet{cuconasu2024pon} show that while random passages unrelated to the question do not affect answer quality, distracting passages, i.e., passages semantically related to the query but not containing the answer, do. Attempts to improve the LLM's robustness mitigate but do not fully solve the LLM's susceptibility to irrelevant content \cite{amiraz2025distracting,jin2025longcontext,linra2024,yoran2024making,yu-etal-2024-chain}.
\citet{amiraz2025distracting} expand this line of research and propose a continuous score to measure the distracting effect of irrelevant passages. Formally, they compute the distracting effect DE$_q(p)$ of an irrelevant passage $p$ for question $q$ as the probability of the LLM not abstaining when receiving only passage $p$ and question $q$ in its prompt:
\noindent
\begin{equation} \label{equ:de}
\text{DE}_q(p) = 1 - p^{\text{LLM}}(\text{NO-RESPONSE}|q, p)
\end{equation}
\noindent
In our paper, we will include this formulation as a core component of our proposed metric for evaluating IR systems serving the RAG scenario. 
A second limitation of the LLM's reading process is the positional bias: the capability of identifying relevant content depends on its location in the prompt. In particular, \citet{liu-etal-2024-lost} discovered the lost-in-the-middle effect, where the LLMs tend to ignore information in the middle of the prompt, and \citet{hutter-2025-lost-but-not-only} demonstrate that different LLMs exhibit distinct positional bias patterns. \citet{cuconasu2025ragsystemssufferpositional} extend this work by showing how the positional bias also affects the impact of distracting documents. 

\subsection{IR Evaluation for RAG}
Most works in the RAG literature evaluate the IR components jointly with the generation components, by assessing the quality of the RAG end-to-end response \cite{Yu_2025}. This approach, in addition to obscuring the retrieval’s distinct contribution, is computationally demanding since it requires generating a response for each new tested context, including contexts that contain the same passages but in a different order. 

When focusing on specifically evaluating the retrieved content, most practitioners still rely on traditional IR metrics, including Precision (Prec), HITS, Mean Average Precision (MAP), Mean Reciprocal Rank (MRR), and normalized Discounted Cumulative Gain (nDCG)\footnote{A detailed description of these metrics is in Appendix \ref{sec:related_metrics}.}. These metrics have been designed for human users, and have relevance as a central notion, which represents the degree to which a retrieved document satisfies a user's information need. Also, they generally apply a discount function to reduce the contribution of documents at lower ranks, reflecting the realistic assumption that users are less likely to examine documents further down in the ranking. 

A few papers propose IR evaluation strategies directly designed for the RAG setting. \citet{salemi2024evaluating} propose eRAG: they score each document based on the quality of the LLM's response when receiving only that document as a context in a RAG setting; then they show that using these scores instead of standard relevance labeling inside traditional IR metrics (e.g., MAP, MRR, nDCG) improves their correlation with the end-to-end RAG performance.
\citet{dai2025seper} propose Semantic Perplexity, a metric that captures the LLM’s internal belief about the correctness of the generated answer, by quantifying the variation of perplexity in the LLM response when answering with and without the retrieved content. The intuition is that useful content leads to increased confidence in the LLM answer.

Similarly to these approaches, in this paper, we employ an LLM-oriented definition of relevance, which we integrate with the notion of distracting effect into a holistic IR metric for the RAG scenario.

\section{Flaws of Traditional IR Metrics in RAG}
\label{sec:pitfalls}

Traditional IR systems and evaluation metrics have been fundamentally designed with human users as the primary consumers of retrieved information, reflecting assumptions about human cognitive processes, information processing capabilities, and search behaviors. These metrics 
inherently assume that users will manually examine ranked result lists, with higher-ranked documents receiving greater attention--a design philosophy rooted in human browsing patterns \cite{10.1145/1008992.1009079,10.1145/1240624.1240690}. 

When the consumers of IR results are LLMs instead of humans, many of these assumptions do not hold.
First, the human-centric notion of document relevance does not necessarily align with the document utility in enabling an LLM to successfully execute its task \cite{tian2025relevancepropagatedretrievergenerator}.  

\begin{table}[htbp]
\small
\centering
\resizebox{\columnwidth}{!}{
\begin{tabular}{@{}lccc@{}}
\toprule
\textbf{Model} & \textbf{Rel in WD} & \textbf{Rel in HD} & \textbf{$\Delta$ (WD - HD)} \\ \midrule
Llama 3B       & 81.11              & 75.56              & 5.55 \\
Llama 8B       & 89.44              & 82.22              & 7.22 \\
Llama 70B      & 90.32              & 83.50              & 6.82 \\
Gemma 4B       & 84.00              & 78.89              & 5.11 \\
Mistral 7B     & 88.33              & 82.00              & 6.33 \\
Qwen 7B        & 87.78              & 78.67              & 9.11 \\ \bottomrule
\end{tabular}
}
\caption{Answer accuracy of LLMs prompted with one relevant passage and four Weak Distractors (WD, DE<0.2) or Hard Distractors (HD, DE>0.8). All differences are statistically significant (Wilcoxon, p < 0.05).}
\label{tab:distraction_pitfall}
\end{table}

Moreover, there is a fundamental difference in how humans and LLMs handle irrelevant documents, as LLMs lack the intuitive filtering mechanisms that allow humans to cleanly separate signal from noise. \citet{jin2025longcontext} provide evidence of this by showing that irrelevant passages retrieved by high-performing retrieval systems are more distracting, and therefore they cause more harm, than irrelevant passages retrieved by weaker retrieval systems. 
Consequently, we argue here that IR systems that are trained and evaluated over traditional IR metrics are likely to optimize for the wrong objective when the consumer of results is an LLM, a machine, rather than a human.

Table \ref{tab:distraction_pitfall} exemplifies this flaw by comparing the average answer accuracy on NQ\footnote{Refer to Section \ref{sec:experimental_setting} for details on LLMs, benchmarks, and answer evaluation.} of several LLMs when prompted with a single relevant passage surrounded by only weakly or only hardly distracting passages. All traditional IR
metrics\footnote{Henceforth, we assume all metrics have a cutoff at $k$.} would score these two scenarios equally, although their end-to-end accuracy changes significantly, exhibiting differences of up to 9 accuracy points.

Another aspect to consider when evaluating IR systems for the RAG scenario is that, unlike humans who sequentially scan through retrieved documents one by one until finding relevant information or abandoning their search, LLMs process the entire prompt holistically, consuming all retrieved documents as a whole. 
Furthermore, the LLMs' ability to leverage the input context is affected by the positional bias. This contradicts the human-centric reward schema characterizing traditional IR metrics such as MRR, nDCG, or MAP, where relevant documents provide a contribution that decreases with their position in the ranking.

\begin{figure}[htp!]
\centering
    \includegraphics[width=7cm]{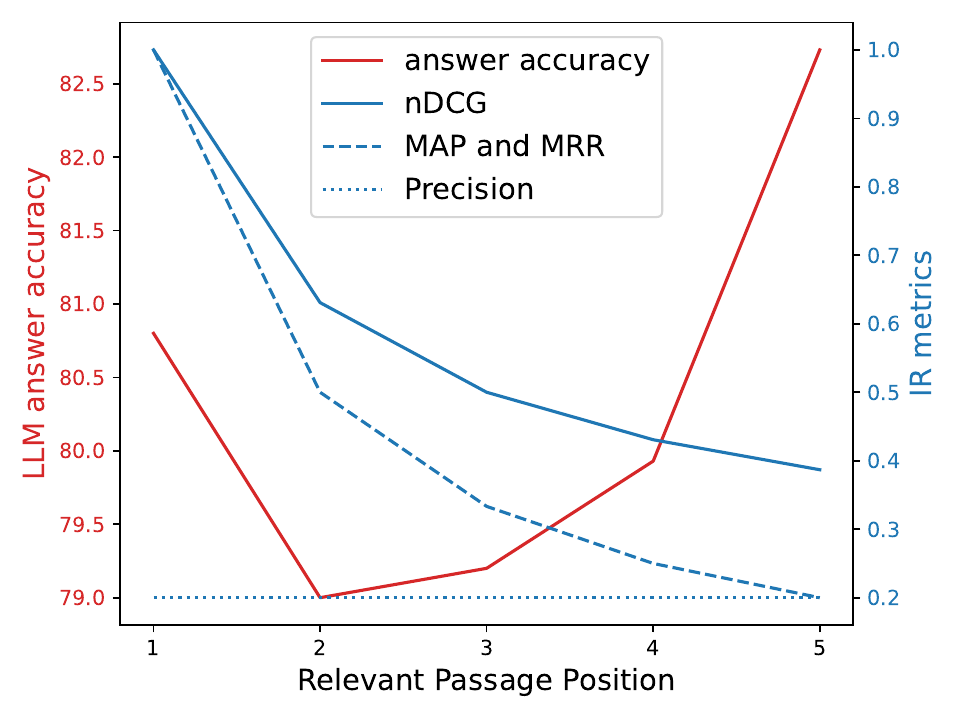}
    \caption{Misalignment between traditional IR metrics (blue) and LLM answer accuracy (red) when a single relevant passage is shifted among irrelevant passages.}
    \label{fig:position_pitfall}
\end{figure}

Figure \ref{fig:position_pitfall} illustrates this 
in a controlled experiment where a single relevant passage is moved in a context of irrelevant ones. Experiments use Qwen 7B on 500 questions from PopQA, NQ, and TriviaQA. LLM accuracy exhibits the U-shaped lost-in-the-middle effect, while nDCG, MAP, and MRR decrease monotonically\footnote{With one relevant passage, MRR equals MAP.}; Precision, being position-unaware, remains constant.

\section{An LLM-oriented Annotation Schema}\label{sec:annotations}
Building on the discussion in the previous section, we propose a novel RAG-oriented annotation schema that replaces the human-centric notion of passage relevance with passage utility for an LLM. Similarly to how the distracting effect is computed, we assess the utility of a passage $p$ for a question $q$ as the probability of the LLM not abstaining when receiving that passage. If the passage is relevant, i.e., if it contains the answer, we interpret this probability as the likelihood that the LLM infers such an answer from the passage\footnote{We assume that relevant passages do not include distracting information. Therefore, when prompted with a lone relevant passage, the LLM either answers correctly or abstains, and never produces wrong answers. Although potentially possible, we never observed this case in preliminary experiments.}. When the passage is irrelevant, the non-abstention probability corresponds to the distracting effect (Eq. \ref{equ:de}) formalized in \citet{amiraz2025distracting}, and we interpret it as a negative utility. 
More formally, given a question $q$ and a passage $p$, the passage utility $u$ is defined as:
\noindent
\begin{equation}\label{eq:utility}
\small
    u(q, p) = R(q, p)  \bigg( 1 - p^{\text{LLM}}(\text{NO-RESPONSE}|q,p) \bigg)
\end{equation}
\noindent 
Where $R(q,p)$ is 1 if the passage $p$ is relevant to the question $q$, -1 otherwise. Such a relevance label can be human-annotated or machine-generated using the LLM-as-a-judge approach.

Note that for a ranked list of $k$ passages, obtaining utility scores for the $k$ individual passages is computationally cheaper than applying an end-to-end evaluation, i.e., running the LLM generator on the entire list and assessing the quality of its response. This is because the runtime cost of a vanilla transformer \cite{NIPS2017_3f5ee243} scales quadratically with its input length. Consequently, if we assume that each passage has an average length of $w$ tokens, and each answer consists on average of $g$ tokens, generating the LLM answer using the full list would cost $\mathcal{O}(gk^2w^2)$; conversely, to obtain the utility of each
passage\footnote{We assume passage relevance annotations are provided (common in IR datasets). Otherwise, they can be computed via LLM-as-a-judge at a similar cost to the abstention probability.}, we would need to invoke $k$ times the LLM with a shorter input and only generate the probability distribution of the first token, for a total cost of $\mathcal{O}(kw^2)$.

\section{Defining a RAG-tailored Metric}\label{sec:metric}
Given a question $q$, the $k$ retrieved passages $\mathcal{C}=[p_1, \ldots, p_k]$ and their utilities $u_i = u(q, p_i)$ to an LLM, we formalize \textbf{Utility and Distraction-aware Cumulative Gain} (UDCG), a metric to score how the entire context $\mathcal{C}$ is expected to help the LLM to answer $q$. We propose two variants of this metric. The former is a learnable version (UDCG$_{\theta}$) that holistically accounts for the concepts of passage relevance, distracting effect, and LLM positional bias and is defined as follows:
\noindent
\begin{equation}\label{eq:full_metric}
    \text{UDCG}_{\theta}(q, \mathcal{C}) = \sigma \bigg (\sum_{i=1}^k\alpha_i u_i^+ + \sum_{i=1}^k\beta_i u_i^- \bigg )
\end{equation}
\noindent 
where $u^+$ and $u^-$ are the positive and negative parts of $u$, respectively, and isolate the relevance and distracting effect of the passages; $\alpha_i$ and $\beta_i$ are positional weights associated with the relevance and the distracting effect of the passages, respectively; finally $\sigma(x) = \frac{1}{1 + e^{-x}}$ is the sigmoid function, which is used to bound the range of the metric in [0, 1] and facilitates the average across multiple questions. 
In short, UDCG$_{\theta}$ calculates a weighted sum of the relevance and the distracting effect of the passages, where the weights take into account the LLM positional bias. 
For contexts having $k$ passages, the metric has $2k$ parameters, i.e., the $\alpha_1, \ldots \alpha_k$ and the $\beta_1, \ldots \beta_k$, which can be learned using a linear model on feature vectors containing $[u^+_1, \ldots u^+_k, u^-_1, \ldots u^-_k]$ trained to predict the end-to-end answer accuracy.

UDCG$_{\theta}$ has two drawbacks: \Ni it requires a training process to learn its $2k$ parameters; and \Nii it strictly depends on the number of retrieved passages $k$, which means that each value of $k$ requires a dedicated training process. To facilitate the metric applicability, we propose its training-free version, named just UDCG for simplicity:
\noindent
\begin{equation}\label{eq:simple_metric}
    \text{UDCG}(q, \mathcal{C}) = \sigma \bigg ( \frac{1}{k}\sum_{i=1}^ku_i^+ + \frac{\gamma}{k}\sum_{i=1}^k u_i^- \bigg )
\end{equation}
\noindent
This version neglects the LLM's positional bias and performs a linear combination of the average passage relevance and the average distracting effect. There is a single hyperparameter $\gamma \in [0,1]$ that balances the contribution of these two terms. It can be tuned for each scenario, but in our experiments, we show that $\gamma=1/3$ provides good results on multiple LLMs and datasets.

\section{Experimental Setting}\label{sec:experimental_setting}

\subsection{Datasets and Models}
We run experiments using five question-answering datasets spanning general knowledge and specialized domains. For general knowledge evaluation, we use PopQA \cite{mallen2023not}, the English subset of NoMIRACL \cite{thakur2024knowing}, and the KILT versions \cite{petroni-etal-2021-kilt} of Natural Questions (NQ) \cite{kwiatkowski2019natural} and TriviaQA \cite{joshi2017triviaqa}. To evaluate performance in specialized domains
, we additionally include the factoid subset of BioASQ \cite{tsatsaronis2015overview} for biomedical question answering. 
For the Wikipedia-based datasets, we use the KILT corpus\footnote{\href{https://huggingface.co/datasets/facebook/kilt_wikipedia}{https://huggingface.co/datasets/facebook/kilt\_wikipedia}}; for BioASQ, we use PubMed\footnote{\href{https://pubmed.ncbi.nlm.nih.gov/}{https://pubmed.ncbi.nlm.nih.gov/}}.
We index the corpora using BGE-large-en-v1.5 embedding model \cite{chen2024bge}
, with the only exception of NoMIRACL, which provides pre-retrieved passages with relevance annotations.
%

As LLMs for answer generation, we use the instruction-tuned versions of Llama-3.2-3B (L-3B), Llama-3.1-8B (L-8B), Llama-3.3-70B (L-70B) \cite{grattafiori2024llama3herdmodels}, Mistral-7B (M-7B) \cite{jiang2023mistral7b}, Gemma-3-4B (G-4B) \cite{gemmateam2025gemma3technicalreport}, and Qwen-2.5-7B (Q-7B) \cite{qwen2025qwen25technicalreport}, spanning different model sizes and families\footnote{For brevity, we omit version numbers when referring to models (e.g., Qwen 7B refers to Qwen-2.5-7B).}. We employ greedy decoding for reproducibility.

\subsection{Evaluation Strategy}
\label{sec:eval_strategy}


Following established practices in RAG evaluation \cite{zheng2023judging, gu2025surveyllmasajudge, rahmani2024report1stworkshoplarge}, we assess passage relevance and answer quality using the LLM-as-a-judge approach. For passage relevance annotation, we use Claude 3.7 Sonnet via AWS Bedrock to classify each retrieved passage as relevant or irrelevant by providing both the query and the reference answer to the model (see Figure \ref{fig:relevance_prompt}). We consider a passage relevant only if it contains the necessary information to answer the query. For NoMIRACL, we utilize the relevance annotations directly provided in the dataset. We then apply our utility-based scoring mechanism to quantify the effects of different passages on LLM effectiveness, as described in Section \ref{sec:annotations}.
%

For the quality evaluation of the responses, we follow previous work on the evaluation of high-fidelity RAG \cite{thakur2024knowing, chen2024aaai} and adopt the setting where LLMs are explicitly asked to abstain when sufficient information is not present in the input context (see Figure \ref{fig:no_res_prompt}).
In this framework, a response is correct only if the input context to the LLM contains at least a relevant passage and its response matches the reference answer (see Figure \ref{fig:answer_eval_prompt}); we evaluate the semantic equivalence of the answer with Gemini 2.0 Flash. 
%
%

This abstention requirement serves multiple purposes: it aligns with real-world deployment scenarios where RAG systems must balance informativeness with reliability, particularly in high-stakes domains (e.g., biomedicine) where incorrect responses can be more harmful than acknowledged uncertainty \cite{sun2025divide}; it 
ensures that responses are only provided when supported by the retrieved content, thereby enhancing faithfulness and user transparency \cite{wallat2025correctness}; and it provides clearer attribution of effectiveness to the retrieval component by preventing LLMs from relying solely on their parametric knowledge when retrieved context is insufficient \cite{xu2024knowledgeconflicts}. 


\section{Experimental Results}

\subsection{Ideal Rankings of Different Annotations}

To show the potential of our annotation schema, we sample 1000 questions from each dataset and retrieve the top 25 passages using BGE-large-en-v1.5. Then we annotate them using \Ni binary relevance labeling, \Nii the eRAG approach \cite{salemi2024evaluating}, which assigns a utility score to each passage by computing the ROUGE-L \cite{lin-2004-rouge} similarity between the reference answer and the answer generated by the LLM when prompted only with that passage, and \Niii our utility-based annotation schema according to Eq. \ref{eq:utility}. For each schema, we assume to have an oracle re-ranker that selects the best five passages based on their annotations\footnote{For traditional relevance labeling, we order the passages based on their binary relevance labels, with relevant passages appearing before irrelevant ones. For passages sharing the same relevance status (either all relevant or all irrelevant), we applied secondary sorting based on their retrieval scores.} and prompt the LLM to generate the response based on such passages.
%
%

%

For 87\% of the queries across the five benchmarks, there is at least a relevant passage among the 25 we retrieved; by selecting passages based on the binary relevance labeling or based on our annotation schema, relevant passages are always ranked better than irrelevant ones; therefore, a perfect LLM would provide 87\% correct responses and abstain in the remaining 13\% of the cases. Conversely, the passage selection according to the eRAG annotation schema sometimes favors irrelevant passages over relevant ones, resulting in only 71\% to 77\% of the contexts that include at least a relevant passage among the selected five, depending on the LLM; this significantly lowers the upper-bound answer correctness achievable with eRAG. 

\begin{table}[hb!]
\centering
\resizebox{\columnwidth}{!}{
\begin{tabular}{@{}l*{9}{c}@{}}
\toprule
 & \multicolumn{3}{c}{\textbf{Binary Rel.}} & \multicolumn{3}{c}{\textbf{eRAG}} & \multicolumn{3}{c}{\textbf{Our Method}} \\
\cmidrule(lr){2-4} \cmidrule(lr){5-7} \cmidrule(l){8-10}
\textbf{Model} & \textcolor{correct}{\textbf{C$\uparrow$}} & \textbf{A} & \textcolor{wrong}{\textbf{W$\downarrow$}} 
 & \textcolor{correct}{\textbf{C$\uparrow$}} & \textbf{A} & \textcolor{wrong}{\textbf{W$\downarrow$}} 
 & \textcolor{correct}{\textbf{C$\uparrow$}} & \textbf{A} & \textcolor{wrong}{\textbf{W$\downarrow$}} \\
\midrule
Llama 3B  & \textcolor{correct}{71.6} & 13.4 & \textcolor{wrong}{15.0} 
          & \textcolor{correct}{60.5} & 23.0 & \textcolor{wrong}{16.5} 
          & \textcolor{correct}{\textbf{72.9}} & 19.0 & \textcolor{wrong}{\textbf{8.2}} \\
Llama 8B  & \textcolor{correct}{73.5} & 15.9 & \textcolor{wrong}{10.6} 
          & \textcolor{correct}{64.4} & 23.6 & \textcolor{wrong}{12.0} 
          & \textcolor{correct}{\textbf{75.1}} & 20.0 & \textcolor{wrong}{\textbf{4.9}} \\
Llama 70B & \textcolor{correct}{78.4} & 12.1 & \textcolor{wrong}{9.5} 
          & \textcolor{correct}{65.9} & 23.7 & \textcolor{wrong}{10.3} 
          & \textcolor{correct}{\textbf{80.2}} & 16.7 & \textcolor{wrong}{\textbf{3.0}} \\
Gemma 4B  & \textcolor{correct}{74.2} & 7.6  & \textcolor{wrong}{18.2} 
          & \textcolor{correct}{66.6} & 11.7 & \textcolor{wrong}{21.7} 
          & \textcolor{correct}{\textbf{76.0}} & 13.1 & \textcolor{wrong}{\textbf{10.9}} \\
Mistral 7B & \textcolor{correct}{76.2} & 10.5 & \textcolor{wrong}{13.3} 
          & \textcolor{correct}{62.7} & 20.8 & \textcolor{wrong}{16.4} 
          & \textcolor{correct}{\textbf{76.9}} & 16.5 & \textcolor{wrong}{\textbf{6.6}} \\
Qwen 7B   & \textcolor{correct}{72.1} & 18.5 & \textcolor{wrong}{9.4} 
          & \textcolor{correct}{62.1} & 26.3 & \textcolor{wrong}{11.5} 
          & \textcolor{correct}{\textbf{73.5}} & 21.8 & \textcolor{wrong}{\textbf{4.8}} \\
\bottomrule
\end{tabular}
}
\caption{Percentage of \textcolor{correct}{correct (C)} answers, abstentions (A), and \textcolor{wrong}{wrong (W)} answers when LLMs are prompted with 5 ideal passages based on Binary Relevance, eRAG, and our utility-based annotations. Results averaged across all tested datasets. All differences are statistically significant (Wilcoxon, p < 0.05).}
\label{tab:oracles_with_abstain}
\end{table}

\begin{table*}[htpb!]
\centering
\resizebox{2.09\columnwidth}{!}{
\begin{tabular}{@{}l*{15}{c}@{}}
\toprule
& \multicolumn{5}{c}{\textbf{Llama 3B}} & \multicolumn{5}{c}{\textbf{Llama 8B}} & \multicolumn{5}{c}{\textbf{Llama 70B}} \\
\cmidrule(lr){2-6} \cmidrule(lr){7-11} \cmidrule(lr){12-16}
\textbf{Method} & \textbf{NQ} & \textbf{Trivia} & \textbf{PopQA} & \textbf{BioASQ} & \textbf{NoMIR} & 
\textbf{NQ} & \textbf{Trivia} & \textbf{PopQA} & \textbf{BioASQ} & \textbf{NoMIR} & 
\textbf{NQ} & \textbf{Trivia} & \textbf{PopQA} & \textbf{BioASQ} & \textbf{NoMIR} \\ 
\midrule
NDCG & 0.458 & 0.545 & 0.438 & 0.618 & 0.326 & 0.515 & 0.623 & 0.499 & 0.699 & 0.332 & 0.560 & 0.708 & 0.669 & 0.752 & 0.509 \\
MRR & 0.452 & 0.543 & 0.436 & 0.617 & 0.315 & 0.510 & 0.619 & 0.501 & 0.698 & 0.323 & 0.560 & 0.704 & 0.667 & 0.751 & 0.503 \\
MAP & 0.454 & 0.543 & 0.438 & 0.617 & 0.320 & 0.513 & 0.620 & 0.499 & 0.697 & 0.329 & 0.558 & 0.707 & 0.667 & 0.751 & 0.506 \\
PREC & 0.482 & 0.600 & 0.462 & 0.631 & 0.373 & 0.546 & 0.671 & 0.532 & 0.714 & 0.369 & 0.591 & 0.734 & 0.682 & 0.769 & 0.539 \\
HITS & 0.411 & 0.559 & 0.444 & 0.590 & 0.358 & 0.509 & 0.638 & 0.514 & 0.716 & 0.359 & 0.570 & 0.741 & 0.662 & 0.804 & 0.550 \\
eRAG & 0.324 & 0.219 & -0.022 & 0.290 & 0.017 & 0.356 & 0.265 & 0.007 & 0.300 & 0.031 & 0.308 & 0.140 & 0.010 & 0.306 & -0.053 \\
\midrule
UDCG & \textbf{0.536} & 0.641 & \textbf{0.525} & 0.665 & 0.482 & 0.611 & 0.711 & 0.599 & 0.718 & 0.503 & \textbf{0.665} & 0.803 & 0.707 & 0.802 & \textbf{0.662} \\
$\text{UDCG}_{\theta}$ & 0.532 & \textbf{0.641} & 0.512 & \textbf{0.668} & \textbf{0.487} & \textbf{0.615} & \textbf{0.713} & \textbf{0.610} & \textbf{0.743} & \textbf{0.503} & 0.661 & \textbf{0.806} & \textbf{0.714} & \textbf{0.811} & 0.653 \\
\midrule
\midrule
& \multicolumn{5}{c}{\textbf{Gemma 4B}} & \multicolumn{5}{c}{\textbf{Mistral 7B}} & \multicolumn{5}{c}{\textbf{Qwen 7B}} \\
\cmidrule(lr){2-6} \cmidrule(lr){7-11} \cmidrule(lr){12-16}
\textbf{Method} & \textbf{NQ} & \textbf{Trivia} & \textbf{PopQA} & \textbf{BioASQ} & \textbf{NoMIR} & 
\textbf{NQ} & \textbf{Trivia} & \textbf{PopQA} & \textbf{BioASQ} & \textbf{NoMIR} & 
\textbf{NQ} & \textbf{Trivia} & \textbf{PopQA} & \textbf{BioASQ} & \textbf{NoMIR} \\ 
\midrule
NDCG & 0.557 & 0.572 & 0.514 & 0.728 & 0.531 & 0.548 & 0.720 & 0.438 & 0.734 & 0.567 & 0.471 & 0.478 & 0.236 & 0.665 & 0.292 \\
MRR & 0.558 & 0.575 & 0.518 & 0.732 & 0.524 & 0.546 & 0.721 & 0.440 & 0.735 & 0.560 & 0.475 & 0.474 & 0.231 & 0.668 & 0.283 \\
MAP & 0.555 & 0.570 & 0.513 & 0.728 & 0.527 & 0.544 & 0.718 & 0.437 & 0.733 & 0.564 & 0.469 & 0.474 & 0.232 & 0.663 & 0.288 \\
PREC & 0.572 & 0.595 & 0.560 & 0.759 & 0.561 & 0.589 & 0.752 & 0.503 & 0.754 & 0.600 & 0.510 & 0.552 & 0.296 & 0.699 & 0.329 \\
HITS & 0.554 & 0.592 & 0.524 & 0.758 & 0.576 & 0.542 & 0.742 & 0.456 & 0.768 & 0.606 & 0.463 & 0.459 & 0.201 & 0.689 & 0.305 \\
eRAG & 0.408 & 0.316 & -0.130 & 0.444 & 0.023 & 0.361 & 0.318 & -0.039 & 0.321 & 0.017 & 0.316 & 0.226 & -0.046 & 0.343 & -0.027 \\
\midrule
UDCG & 0.587 & 0.681 & \textbf{0.607} & 0.776 & \textbf{0.626} & 0.629 & \textbf{0.786} & 0.542 & 0.767 & 0.609 & \textbf{0.605} & 0.649 & 0.397 & 0.752 & \textbf{0.562} \\
$\text{UDCG}_{\theta}$ & \textbf{0.593} & \textbf{0.682} & 0.598 & \textbf{0.793} & 0.618 & \textbf{0.634} & 0.783 & \textbf{0.556} & \textbf{0.777} & \textbf{0.632} & 0.601 & \textbf{0.650} & \textbf{0.406} & \textbf{0.755} & 0.550 \\
\bottomrule
\end{tabular}
}
\caption{Spearman correlation between IR metrics and end-to-end RAG accuracy. Bold indicates best per model-dataset combination. UDCG improvements are statistically significant (bootstrap, p < 0.05).}
\label{tab:main_results}
\end{table*}

Table \ref{tab:oracles_with_abstain} reports the average results across the five benchmarks.
Our annotation schema always improves results significantly, demonstrating that achieving perfect traditional IR metrics is insufficient for optimal RAG performance, calling for the development of new evaluation frameworks that better align with the objectives of modern retrieval-augmented generation systems. In particular, our annotation schema, by selecting all possible relevant passages and completing the context with the irrelevant ones that minimize the distracting effect, facilitates the LLM task of inferring the right answer when present in the passage context, leading to up to 2\% more correct answers than the binary relevance labeling. Additionally, when no relevant passage is present, the absence of hard distracting passages helps the LLM to abstain, reducing by half the wrong answers.
The improvement w.r.t. eRAG is even more evident, and we argue that the main reasons for this difference are \Ni eRAG neglects the damaging impact of distracting documents, and \Nii eRAG uses ROUGE-L, which is a weak estimator of the answer quality, as other token-level answer equivalence measures \cite{bulian-etal-2022-tomayto}.

\subsection{Correlation between IR Metrics and Answer Accuracy}
\label{sec:main_res}




To validate the effectiveness of our proposed metrics, we conducted comprehensive experiments examining their correlation with the end-to-end answer accuracy of the RAG system. Specifically, given a question $q$, we create 10 different contexts $\mathcal{C}_1, \dots, \mathcal{C}_{10}$, each one containing 5 passages, which we use to generate the corresponding LLM answers $a_1, \ldots, a_{10}$. Then we assess the correctness of each answer and sort the contexts so that contexts leading to correct answers are positioned first, followed by contexts leading to abstation, followed by contexts leading to wrong answers. We compare this ideal context ranking with the one induced by a metric and compute their Spearman correlation.
We create contexts by randomly sampling passages from the top-25 retrieved ones. We balanced the resulting examples so that 50\% of the contexts contain at least a relevant passage, while the remaining 50\% contain only irrelevant passages. 

To learn the $\theta = \{\alpha_1, \ldots \alpha_k, \beta_1, \ldots \beta_k \}$ parameters of UDCG$_{\theta}$, we use a linear SVM$^{rank}$ model \cite{10.1145/775047.775067} with default hyperparameters\footnote{\url{https://www.cs.cornell.edu/people/tj/svm_light/svm_rank.html}}. As a training set, we use 1200 query-context examples derived from NQ, and tested on 500 query-context examples extracted from a disjoint set of NQ questions. Furthermore, we assess whether the learned metric generalizes to other datasets, by testing on 500 query-context samples of TriviaQA, PopQA, NoMIRACL, and BioASQ. 


Table \ref{tab:main_results} reports the results of different LLMs. Our proposed metrics, UDCG$_{\theta}$ and its training-free version UDCG, consistently achieve the best correlation with the end-to-end RAG accuracy, across all LLMs and benchmarks, with UDCG$_\theta$ achieving a 36\% (+10pts) increase in terms of Spearman correlation when compared to NDCG (averaging across all considered LLMs and datasets). UDCG$_{\theta}$, although trained only on NQ, maintains competitive results on all benchmarks, demonstrating its generalized applicability. 

\begin{table}[ht!]
\centering
\resizebox{\columnwidth}{!}{
\begin{tabular}{@{}lcccccc@{}}
\toprule
\textbf{Metric} & \textbf{L-3B} & \textbf{L-8B} & \textbf{L-70B} & \textbf{G-4B} & \textbf{M-7B} & \textbf{Q-7B} \\ \midrule
UDCG$_\theta$         & \textbf{0.532} & \textbf{0.615} & \textbf{0.661} & \textbf{0.593} & \textbf{0.634} & \textbf{0.601} \\
~~(rel-only)          & 0.508          & 0.594          & 0.625          & 0.567          & 0.607          & 0.553 \\
~~(binary)            & 0.473          & 0.542          & 0.589          & 0.554          & 0.584          & 0.498 \\
\midrule
UDCG                  & \textbf{0.536} & \textbf{0.611} & \textbf{0.665} & \textbf{0.587} & \textbf{0.629} & \textbf{0.605} \\
~~(rel-only)          & 0.515          & 0.599          & 0.638          & 0.572          & 0.606          & 0.553 \\
\bottomrule
\end{tabular}
}
\caption{Ablation study. Spearman correlation with RAG accuracy. UDCG consistently outperforms relevance-only baselines.}
\label{tab:ablation}
\end{table}

Surprisingly, the difference between UDCG and UDCG$_{\theta}$ is minor; conversely to UDCG, UDCG$_{\theta}$ accounts for the passage positions in the prompt, but we argue that this information is not very important to estimate the end-to-end accuracy. Indeed, as observed by \citet{cuconasu2025ragsystemssufferpositional}, the effect of positional bias, which is often amplified in controlled settings, is marginal in real scenarios.

\subsection{Ablation Study for UDCG}

We examine the contribution of each component in our proposed metrics through an ablation study on the NQ dataset across all LLMs.
Table~\ref{tab:ablation} reports the Spearman correlation between metric scores and end-to-end RAG accuracy under different configurations.
For UDCG$_{\theta}$, we train three variants independently: the full model (see Eq.~\ref{eq:full_metric}) incorporating both relevance utility features ($u_i^+$) and distracting effect features ($u_i^-$); a relevance-only version, where we disable the distracting effect features; and a binary version where we additionally constrain the relevance utility features $u_i^+=1$ to be binary values reflecting the traditional binary relevance labeling. 
%
%
Removing the distracting effect (rel-only) causes consistent degradation across all models, with correlation drops ranging from 3.4\% to 8\%. The binary variant shows even larger degradation, with drops of up to 17\%, highlighting the importance of our fine-grained utility-based annotations over binary relevance judgments.
For the training-free version of UDCG, we compare the full model (see Eq.~\ref{eq:simple_metric}) against a relevance-only version ($\gamma = 0$), which completely removes the contribution of distracting
passages\footnote{Note that further substituting continuous utility with binary relevance yields Precision (see Table \ref{tab:main_results}).}. 
Consistent with UDCG$_{\theta}$, removing distracting effects leads to an average correlation decrease of 4\%.
These results confirm that our utility-based scoring, which includes the distracting effects, is essential for accurately predicting RAG system effectiveness.

\subsection{Sensitivity to Context Length}
\begin{figure}[t!]
\centering
    \includegraphics[width=7.5cm]{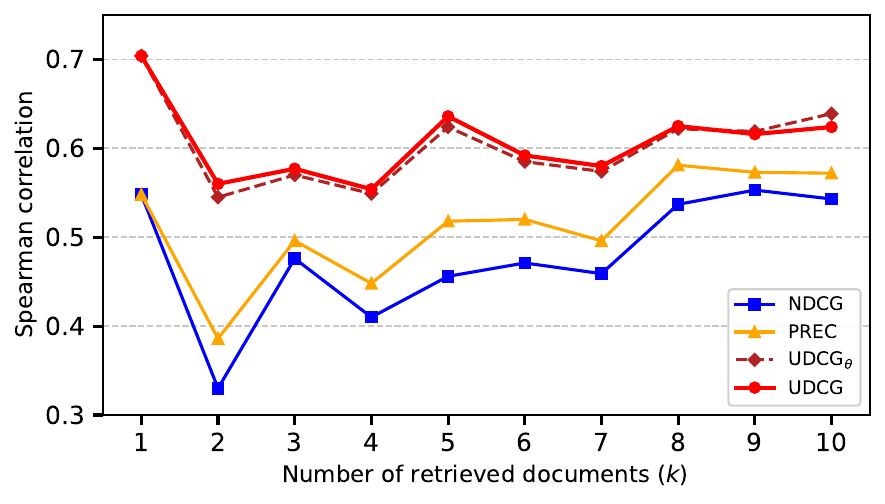}
    \caption{IR metric robustness across context sizes for Qwen 7B. UDCG maintains a stable correlation with RAG accuracy for all $k$.}
\label{fig:abl_k}
\end{figure}

We evaluate whether our proposed metrics maintain consistent predictive power as the number of retrieved documents k varies. We conduct this study using Qwen 7B on NQ, systematically varying $k$ from 1 to 10 passages. We limit our analysis to $k \leq 10$ because additional documents at lower ranks would likely consist of distractors with a negative impact on the generation quality \cite{cuconasu2025ragsystemssufferpositional}.
%
Figure \ref{fig:abl_k} presents the results. UDCG demonstrates remarkable stability across different context sizes, maintaining correlations between 0.554 and 0.704, while consistently outperforming all baseline metrics across all values of $k$. In contrast, traditional IR metrics show substantially more volatility, with NDCG and Precision exhibiting standard deviations that are respectively 1.6 and 1.38 times higher than UDCG\footnote{UDCG$_\theta$ shows similar stability but requires training for each $k$, making it less practical for dynamic context sizes.}.

\section{Conclusions}


This paper addresses a fundamental misalignment between traditional information retrieval evaluation metrics and the requirements of modern RAG systems. We demonstrated that classical IR metrics like nDCG, MAP, and MRR fail to adequately predict RAG performance because they assume monotonically decreasing document utility with rank position and ignore the heterogeneous distracting effects of irrelevant documents on LLM generation quality.
To address these limitations, we proposed a novel RAG-oriented annotation schema that replaces the human-centric notion of passage relevance with positive or negative passage utility for an LLM. 
Building on this foundation, we defined UDCG, a metric using an LLM-oriented positional discount to directly optimize the correlation with the end-to-end answer accuracy.
Notably, we have shown that removing positional discounting entirely achieves nearly identical performance.
Our extensive experiments on five QA datasets and six LLMs demonstrate that both metrics consistently outperform traditional IR metrics and existing RAG-oriented approaches in predicting end-to-end system performance by up to 36\%.

%
%
%

%

\section*{Limitations}

Our work has some limitations that suggest directions for future research. First, our experiments focus on the question-answering task where queries have specific and verifiable answers. The applicability of our metrics to other tasks, such as multi-hop question answering or fact verification, remains unexplored.
Second, our utility-based annotation schema requires access to the LLM's output probabilities to compute abstention likelihood. This requirement limits our approach to settings where logit access is available, excluding black-box commercial APIs that only return generated text. 
Third, our evaluation is conducted exclusively on English-language datasets. The effectiveness of our utility-based scoring and the UDCG metric in multilingual or cross-lingual RAG settings remains to be investigated, as language-specific characteristics may influence both passage utility and distracting effects.

\bibliography{main}


\appendix

\section{Traditional IR Metrics}\label{sec:related_metrics}

The offline evaluation of information retrieval systems relies heavily on a well-established set of metrics that measure different aspects of retrieval performance. These traditional metrics have relevance as a central notion, which represents the degree to which a retrieved document satisfies a user's information need. The conventional approach to relevance in IR has been predominantly binary, where documents are classified as either relevant or non-relevant to a given query. The IR metrics proposed in the literature mostly differ in how they account for the presence and positioning of relevant documents in the retrieved document rank.

Precision and Recall serve as the foundational metrics in IR evaluation. Precision measures the fraction of retrieved documents that are relevant, while recall measures the fraction of relevant documents that are successfully retrieved. For modern (web-scale) information retrieval, recall is no longer a meaningful metric, as many queries have thousands of relevant documents, and few users will be interested in reading all of them. Precision@$k$ ($P@k$), by measuring precision at fixed cutoff points, reflects practical constraints where users typically examine only the top-$k$ results.

Mean Average Precision (MAP) extends previous metrics by incorporating ranking information. MAP computes the average precision at each relevant document position and then averages across all queries in a test collection. This metric effectively rewards systems that rank relevant documents higher in the result list, making it particularly suitable for evaluating systems where users examine results sequentially. More formally:
\noindent
\begin{equation*}
    \text{MAP} = \frac{1}{|Q|} \sum_{q \in Q} \frac{1}{R_q} \sum_{k=1}^n P@k \times r(k)
\end{equation*}

\noindent where $Q$ is the set of test queries, $R_q$ is the number of existing relevant documents for query $q$, $n$ is the number of retrieved documents, and $r(k)$ is an indicator function equaling 1 if the item at rank $k$ is a relevant document, zero otherwise.

Mean Reciprocal Rank (MRR) focuses specifically on the rank position of the first relevant document, making it particularly appropriate for navigational queries where users seek a single correct answer. MRR computes the reciprocal of the rank at which the first relevant document appears, namely rank$_i$, and averages this value across all queries:

\begin{equation*}
    \text{MRR} = \frac{1}{|Q|}\sum_{q \in Q} \frac{1}{\text{rank}_i}
\end{equation*}


Normalized Discounted Cumulative Gain (nDCG) enables the usage of graded relevance judgments rather than binary values. nDCG recognizes that documents can have varying degrees of relevance and that highly relevant documents should contribute more to the overall evaluation score. The metric applies a logarithmic discount function to reduce the contribution of documents at lower ranks, reflecting the realistic assumption that users are less likely to examine documents further down in the ranking. The DCG accumulated at a particular rank position $p$ is defined as:
\noindent
\begin{equation*}
    \text{DCG}_p = \sum_{i=1}^p \frac{rel_i}{\log_2(i+1)}
\end{equation*}

\noindent where $rel_i$ is the graded relevance of the result at position $i$ retrieved for the query $q$. This value is typically normalized to facilitate comparison across queries with different numbers of relevant documents. To this end, we sort documents of a result list by relevance, producing an ideal DCG at position $p$ (IDCG$_p$), and use this value as a normalization factor: 
\noindent
\begin{equation*}
    \text{nDCG}_p = \frac{\text{DCG}_p}{\text{IDCG}_p}
\end{equation*}

The nDCG values for all queries can be averaged to obtain a measure of the average performance of a search engine's ranking algorithm.

Finally, HITS measures whether there is at least a relevant passage in the top-$k$ retrieved ones.

\section{Open-source Alternative for Relevance}

In Section \ref{sec:eval_strategy}, we describe our method to compute the relevance of a passage with an LLM-as-a-Judge approach. In our analysis, we used Claude 3.7 Sonnet; however, we also tested Llama-3.3-70B Instruct as a strong open-source alternative for our evaluation tasks. 
We sampled 300 queries from our datasets and evaluated relevance decisions for the top-5 passages retrieved by BGE using both models, analyzing a total of 1500 passages. For passage relevance assessment, the Cohen's Kappa coefficient between Claude and Llama yielded a score of 0.85, indicating substantial agreement according to standard interpretation guidelines \cite{landis1977measurement}.
These results suggest that Llama-3.3-70B Instruct can serve as a reliable open-source alternative for passage relevance assessment in RAG experiments.

\begin{figure*}[b]
\begin{mdframed}[font=\footnotesize]
\begin{Verbatim}[breaklines=true, breaksymbol=]
You are given a question and you must respond based on the provided documents. Respond directly without providing any premise or explanation. If none of the documents contain the answer, please respond with NO-RESPONSE. Do not try to respond based on your own knowledge.

Documents:
<document>

Question: 
<question>

Answer:
\end{Verbatim}
\end{mdframed}
\vspace{-0.4cm}
\caption{Prompt for evaluating the utility of a document and for answer generation.} \label{fig:no_res_prompt}
\end{figure*}

\begin{figure*}[b]
\begin{mdframed}[font=\footnotesize]
\begin{Verbatim}[breaklines=true, breaksymbol=]
Determine if a document is RELEVANT or IRRELEVANT for answering a question. A document is RELEVANT if it contains information that directly supports at least one acceptable answer.

RELEVANT examples:
- Q: "where does the story the great gatsby take place" | Answers: ['Long Island of 1922']
  Doc: "The Great Gatsby...follows characters living in West Egg on Long Island in summer of 1922"
  → Contains the exact answer

- Q: "when did korn's follow the leader come out" | Answers: ['August 18, 1998', 'Summer 1998']
  Doc: "Follow the Leader...was released on August 18, 1998"
  → Contains the exact release date

IRRELEVANT examples:
- Q: "where does the story the great gatsby take place" | Answers: ['Long Island of 1922']
  Doc: "While Long Island features prominently in American literature, the socioeconomic dynamics..."
  → Mentions Long Island but not as the story's setting

- Q: "who played bobby byrd in get on up" | Answers: ['Nelsan Ellis']
  Doc: "Critics praised the casting of Bobby Byrd and the chemistry between the main characters...
  → Discusses casting but doesn't name the specific actor

Evaluation steps:
1. Find the specific information needed to match an acceptable answer. Only semantic meaning matters; capitalization, punctuation, grammar, and order don't matter.
2. Check if the document contains this information directly or through clear inference
3. Check for these common errors:
    - The document contains similar keywords/themes but not the actual answer
    - The document contains partial information that would need to be combined with external knowledge
    - The document discusses related topics but doesn't specifically answer the question

Here is a new example. Don't apologize or correct yourself if there was a mistake; we are just trying to evaluate the relevance of the document.
```
Question: {question}
Acceptable answers list: {answers}
Document: {document}
```

Evaluate the document for this new question as one of:
A: RELEVANT
B: IRRELEVANT

Return a JSON object with the following format:
{
    "motivation": "Brief explanation (max 2 sentences)",
    "grade": "A" or "B"
}
\end{Verbatim}
\end{mdframed}
\vspace{-0.4cm}
\caption{Prompt template for document relevance assessment using Claude 3.7 Sonnet as judge.} \label{fig:relevance_prompt}
\end{figure*}

\begin{figure*}[b]
\begin{mdframed}[font=\scriptsize]
\begin{Verbatim}[breaklines=true, breaksymbol=]
Your job is to grade a predicted answer as CORRECT, INCORRECT, or ABSTAINED against a list of acceptable answers.

## Grading Rules

A predicted answer is CORRECT if:
- It contains the essential information from at least one acceptable answer
- It contains no contradictory information
- When the acceptable answer is a list, at least one item from that list is present

A predicted answer is INCORRECT if:
- It contradicts any acceptable answer
- It contains factually wrong information (even with hedging like "I think" or "possibly")

A predicted answer is ABSTAINED if:
- It explicitly refuses to answer or states uncertainty without providing substantive information
- It contains phrases like "I don't know", "I'm not sure", "I cannot answer", "NO-RESPONSE", "I'm uncertain"

## Examples

**CORRECT answers:**
```
Question: What are Barack Obama's children's names?
Acceptable: ['Malia Obama and Sasha Obama', 'Natasha Marian and Malia Ann']
"Malia and Sasha" (essential info present)
"I believe it's Malia and Sasha Obama" (hedging with correct info)
```

**INCORRECT answers:**
```
Question: What are Barack Obama's children's names?
Acceptable: ['Malia Obama and Sasha Obama']
"Only Malia" (incomplete when both required)
"Malia, Sasha, and Susan" (contains wrong info)
"He has no children" (contradicts acceptable answer)
```

**ABSTAINED answers:**
```
Question: What are Barack Obama's children's names?
"I don't know the names of his children"
"I'm not sure about this information"
"I cannot provide that information"
"I'm uncertain and don't want to guess"
"NO-RESPONSE"
```

## Important Notes
- Ignore capitalization, punctuation, grammar, and order.
- Don't penalize omitted information that's clearly inferred from the question.
- Don't penalize minor typos in names.
- Grade using only the question and acceptable answers provided, even if you know the answer is factually wrong.

---

Here is a new example. Don't apologize or correct yourself if there was a mistake; we are just trying to grade the answer.
```
Question: {question}
Acceptable answers list: {target}
Predicted answer: {predicted_answer}
```

Grade the predicted answer of this new question as one of:
A: CORRECT
B: INCORRECT
C: ABSTAINED

Return a JSON object with the following format:
{
 "motivation": "Your concise motivation for the grade here. Use maximum 2 sentences.",
 "grade": "A", "B", or "C"
}
\end{Verbatim}
\end{mdframed}
\vspace{-0.4cm}
\caption{Prompt template for answer correctness assessment using Gemini 2.0 Flash as judge.} \label{fig:answer_eval_prompt}
\end{figure*}

\end{document}